\pdfoutput=1

\documentclass[11pt]{article}

\usepackage{acl}

\usepackage{times}
\usepackage{latexsym}

\usepackage[T1]{fontenc}

\usepackage[utf8]{inputenc}

\usepackage{microtype}
\usepackage{graphicx}
\graphicspath{Figures}

\usepackage{xcolor}	

\usepackage{times}
\usepackage{latexsym}
\usepackage{arydshln}
\usepackage{graphicx}
\usepackage{subcaption}
\usepackage{hyperref}
\usepackage{lscape}
\usepackage{multirow}
\usepackage{booktabs}
\usepackage{amsmath}

\DeclareMathOperator*{\argsort}{arg\,sort}

\usepackage{algorithm}
\usepackage{algpseudocode}

\usepackage{microtype,enumitem}
\usepackage{xcolor}	

\usepackage[export]{adjustbox}
\usepackage[utf8]{inputenc}

\usepackage{float}
\newfloat{algorithm}{t}{lop}
\usepackage{adjustbox}

\title{Topic-Guided Sampling For Data-Efficient Multi-Domain Stance Detection}

\author{Erik Arakelyan$^{1}$, Arnav Arora$^{2}$, Isabelle Augenstein$^{3}$ \\
Department of Computer Science\\
University of Copenhagen \\
Copenhagen Denmark  \\ 
\texttt{\{erik.a,aar,augenstein\}@di.ku.dk} 
}

\begin{document}
\maketitle
\begin{abstract}

Stance Detection is concerned with identifying the attitudes expressed by an author towards a target of interest. This task spans a variety of domains ranging from social media opinion identification to detecting the stance for a legal claim. However, the framing of the task varies within these domains, in terms of the data collection protocol, the label dictionary and the number of available annotations. Furthermore, these stance annotations are significantly imbalanced on a per-topic and inter-topic basis. These make multi-domain stance detection a challenging task, requiring standardization and domain adaptation. To overcome this challenge, we propose \textbf{T}opic \textbf{E}fficient \textbf{St}anc\textbf{E} \textbf{D}etection	(TESTED), consisting of a topic-guided diversity sampling technique and a contrastive objective that is used for fine-tuning a stance classifier. We evaluate the method on an existing benchmark of $16$ datasets with in-domain, i.e. all topics seen and out-of-domain, i.e. unseen topics, experiments. The results show that our method outperforms the state-of-the-art with an average of $3.5$ F1 points increase in-domain, and is more generalizable with an averaged increase of $10.2$ F1 on out-of-domain evaluation while using $\leq10\%$ of the training data. We show that our sampling technique  mitigates both inter- and per-topic class imbalances. Finally, our analysis demonstrates that the contrastive learning objective allows the model a more pronounced segmentation of samples with varying labels.


\end{abstract}

\section{Introduction}
\label{sec:intro}

The goal of stance detection is to identify the viewpoint expressed by an author within a piece of text towards a designated topic \citep{mohammad2016semeval}. Such analyses can be used in a variety of domains ranging from identifying claims within political or ideological debates~\citep{somasundaran2010recognizing, thomas2006get}, identifying mis- and disinformation \citep{hanselowski2018retrospective,hardalov2021survey}, public health policymaking \citep{glandt2021stance,hossain2020covidlies, osnabrugge2023cross}, news recommendation \citep{reuver2021no} to investigating attitudes voiced on social media  \citep{qazvinian2011rumor,augenstein-etal-2016-stance,conforti2020will}. However, in most domains, and even more so for cross-domain stance detection, the exact formalisation of the task gets blurry, with varying label sets and their corresponding definitions, data collection protocols and available annotations. Furthermore, this is accompanied by significant changes in the topic-specific vocabulary \citep{somasundaran2010recognizing,wei2019modeling}, text style \citep{pomerleau2017fake, ferreira2016emergent} and topics mentioned either explicitly \citep{qazvinian2011rumor, walker2012corpus} or implicitly \citep{hasan2013stance,gorrell2019semeval}. Recently, a benchmark of $16$ datasets \citep{hardalov2021cross} covering a variety of domains and topics has been proposed for testing stance detection models across multiple domains. It must be noted that these datasets are highly imbalanced, with an imbalanced label distribution between the covered topics, i.e. inter-topic and within each topic, i.e. per-topic, as can be seen in \autoref{fig:imbalanced} and \autoref{fig:label_dist}. This further complicates the creation of a robust stance detection classifier.

\begin{figure}[t]
\centering
\includegraphics[width=\columnwidth]{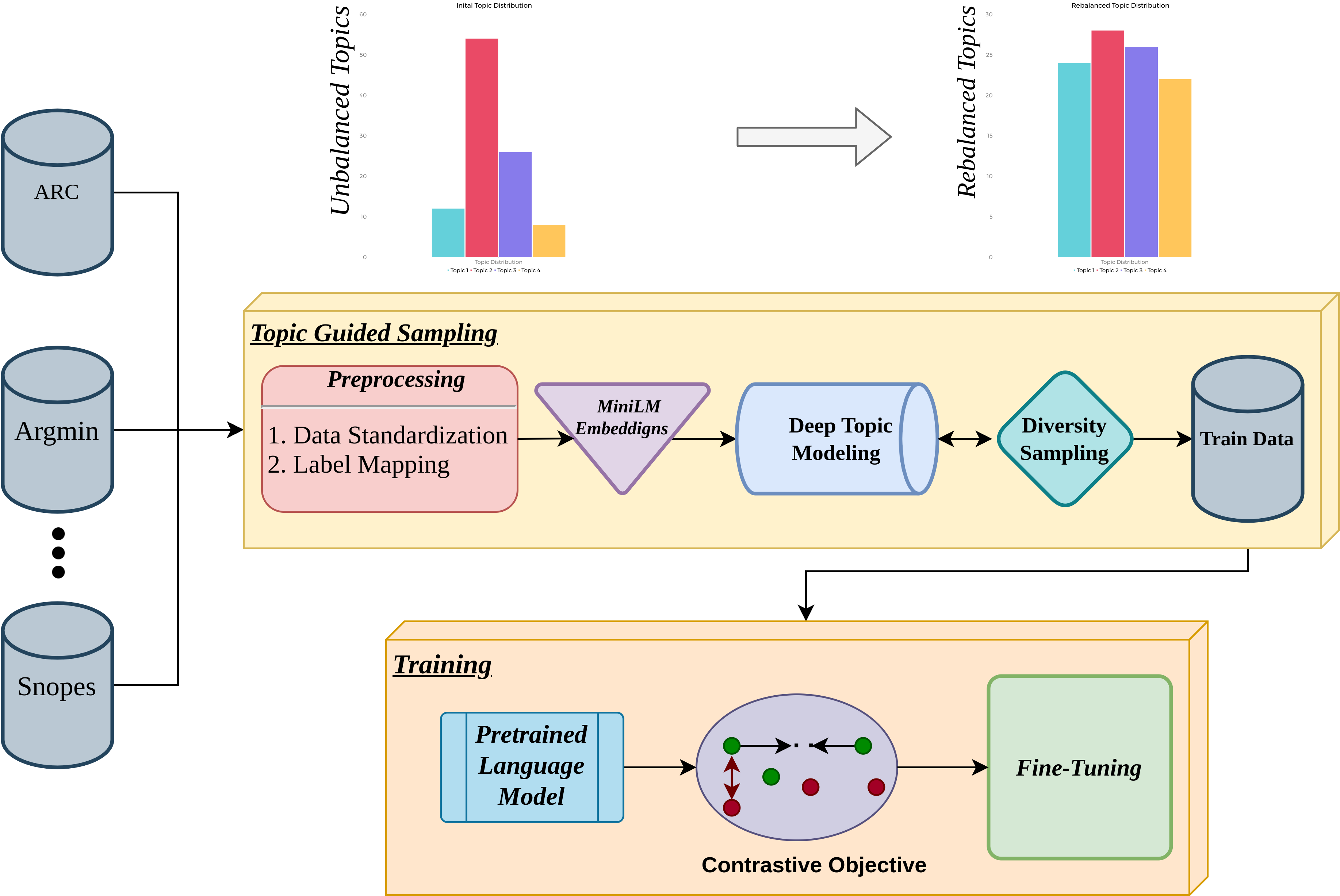}
\caption{The two components of TESTED: Topic Guided Sampling (top) and training with contrastive objective (bottom).}
\label{fig:TESTED}
\end{figure}



Given the inherent skew present within the dataset and variances within each domain, we propose a topic-guided diversity sampling method, which produces a data-efficient representative subset while mitigating label imbalances. These samples are used for fine-tuning a Pre-trained Language Model (PLM), using a contrastive learning objective to create a robust stance detection model. These two components form our \textbf{T}opic \textbf{E}fficient \textbf{St}anc\textbf{E} \textbf{D}etection	(TESTED) framework, as seen in \autoref{fig:TESTED}, and are analysed separately to pinpoint the factors impacting model performance and robustness. We test our method on the multi-domain stance detection benchmark by \citet{hardalov2021cross}, achieving state-of-the-art results with both in-domain, i.e. all topics seen and out-of-domain, i.e. unseen topics evaluations. Note though that TESTED could be applied to any text classification setting.

In summary, our \textbf{contributions} are:
\begin{itemize}[noitemsep]
    \item We propose a novel framework (TESTED) for predicting stances across various domains, with data-efficient sampling and contrastive learning objective;
    \item Our proposed method achieves SOTA results both in-domain and out-of-domain;
    \item Our analysis shows that our topic-guided sampling method mitigates dataset imbalances while accounting for better performance than other sampling techniques; 
    \item The analysis shows that the contrastive learning objective boosts the ability of the classifier to differentiate varying topics and stances.
\end{itemize}

\section{Related Work}
\label{sec:related_work}

\paragraph{Stance Detection} is an NLP task which aims to identify an author's attitude towards a particular topic or claim. The task has been widely explored in the context of mis- and disinformation detection \citep{ferreira2016emergent,hanselowski2018retrospective,ZUBIAGA2018273,hardalov2021survey}, sentiment analysis \citep{mohammad2017stance, aldayel2019your} and argument mining \citep{boltuvzic2014back,sobhani2015argumentation,wang2019survey}. Most papers formally define stance detection as a pairwise sequence classification where stance targets are provided \citep{kuccuk2020stance}. However, with the emergence of different data sources, ranging from debating platforms \citep{somasundaran2010recognizing,hasan2014you,aharoni2014benchmark} to social media \citep{mohammad2016semeval,gorrell2019semeval}, and new applications \citep{zubiaga2018detection,hardalov2021survey}, this formal definition has been subject to variations w.r.t. the label dictionary inferred for the task.

Previous research has predominantly focused on a specific dataset or domain of interest, outside of a few exceptions like multi-target \citep{sobhani2017dataset,wei2018multi} and cross-lingual \citep{hardalov2022few} stance detection. In contrast, our work focuses on multi-domain stance detection, while evaluating in- and out-of-domain on a $16$ dataset benchmark with state-of-the-art baselines \citep{hardalov2021cross}.

\paragraph{Topic Sampling}

Our line of research is closely associated with diversity \citep{ren2021survey} and importance \citep{beygelzimer2009importance} sampling and their applications in natural language processing \citep{zhu2008active,zhou2020informed}. Clustering-based sampling approaches have been used for  automatic speech recognition \citep{syed2016supervised}, image classification \citep{ranganathan2017deep,yan2022mitigating} and semi-supervised active learning \citep{buchert2022exploiting} with limited use for textual data \citep{yang2014active} through topic modelling \citep{blei2003latent}. This research proposes an importance-weighted topic-guided diversity sampling method that utilises deep topic models, for mitigating inherent imbalances present in the data, while preserving relevant examples.

\paragraph{Contrastive Learning}  has been used for tasks where the expected feature representations should be able to differentiate between similar and divergent inputs \citep{liu2021self,10.1145/3561970}. Such methods have been used for image classification \citep{khosla2020supervised}, captioning \citep{dai2017contrastive} and textual representations \citep{giorgi2020declutr,jaiswal2020survey,ostendorff-etal-2022-neighborhood}. The diversity of topics \citep{qazvinian2011rumor,walker2012corpus,hasan2013stance}, vocabulary \citep{somasundaran2010recognizing,wei2019modeling} and expression styles \citep{pomerleau2017fake} common for stance detection can be tackled with contrastive objectives, as seen for similar sentence embedding and classification tasks \citep{gao2021simcse, yan2021consert}. 

\section{Datasets}
\label{sec:datasets}
Our study uses an existing multi-domain dataset benchmark \citep{hardalov2021cross}, consisting of $16$ individual datasets split into four source groups: \textit{Debates, News, Social Media, Various}. The categories include datasets about debating and political claims including arc \citep{hanselowski2018retrospective,habernal2017argument}, iac1 \citep{walker2012corpus}, perspectum \citep{chen2019seeing}, poldeb \citep{somasundaran2010recognizing}, scd \citep{hasan2013stance}, news like emergent \citep{ferreira2016emergent}, fnc1 \citep{pomerleau2017fake}, snopes \citep{hanselowski-etal-2019-richly}, social media like mtsd \cite{sobhani2017dataset}, rumour \citep{qazvinian2011rumor}, semeval2016t6 \citep{mohammad2016semeval}, semeval2019t7 \citep{gorrell2019semeval}, wtwt \citep{conforti2020will} and datasets that cover a variety of diverse topics like argmin \citep{stab-etal-2018-cross}, ibmcs \citep{bar-haim_stance_2017} and vast \citep{allaway-mckeown-2020-zero}. Overall statistics for all of the datasets can be seen in \autoref{Appendix:data}.

\subsection{Data Standardisation}

As the above-mentioned stance datasets from different domains possess different label inventories, the stance detection benchmark by \citet{hardalov2021cross} introduce a mapping strategy to make the class inventory homogeneous. We adopt that same mapping for a fair comparison with prior work, shown in Appendix \ref{Appendix:data}.

\section{Methods}

Our goal is to create a stance detection method that performs strongly on the topics known during training and can generalize to unseen topics. The benchmark by \citet{hardalov2021cross} consisting of $16$ datasets is highly imbalanced w.r.t the inter-topic frequency and per-topic label distribution, as seen in \autoref{fig:imbalanced}. 

These limitations necessitate a novel experimental pipeline. The first component of the pipeline we propose is an importance-weighted topic-guided diversity sampling method that allows the creation of supervised training sets while mitigating the inherent imbalances in the data. We then create a stance detection model by fine-tuning a Pre-trained Language Model (PLM) using a contrastive objective.


\label{sec:methods}

\subsection{Topic-Efficient Sampling}
\label{sec:methods:topic}

We follow the setting in prior work on data-efficient sampling \cite{buchert2022exploiting,yan2022mitigating}, framing the task as a selection process between multi-domain examples w.r.t the theme discussed within the text and its stance. This means that given a set of datasets $\mathcal{D} = (\mathcal{D}_1, \dots \mathcal{D}_n)$ with their designated documents $\mathcal{D}_i = (d_{i}^1, \dots d_{i}^m)$, we wish to select a set of diverse representative examples $\mathcal{D}_{\textbf{train}}$, that are balanced w.r.t the provided topics $\mathcal{T} = (t_1, \dots t_q)$ and stance labels $L = (l_1, \dots l_k)$. 


\paragraph{Diversity Sampling via Topic Modeling}

We thus opt for using topic modelling to produce a supervised subset from all multi-domain datasets. Selecting annotated examples during task-specific fine-tuning is a challenging task \cite{shao2019learning}, explored extensively within active learning research \cite{hino2020active, konyushkova2017learning}. Random sampling can lead to poor generalization and knowledge transfer within the novel problem domain \cite{das2021importance, perez2021true}. To mitigate the inconsistency caused by choosing suboptimal examples, we propose using deep unsupervised topic models, which allow us to sample relevant examples for each topic of interest. We further enhance the model with an importance-weighted diverse example selection process \cite{shao2019learning,yang2015multi} within the relevant examples generated by the topic model. The diversity maximisation sampling is modeled similarly to \citet{yang2015multi}.

The topic model we train is based on the technique proposed by \citet{angelov2020top2vec} that tries to find topic vectors while jointly learning document and word semantic embeddings. The topic model is initialized with weights from the \textit{all-MiniLM-L6} PLM, which has a strong performance on sentence embedding benchmarks \citep{wang2020minilm}. It is shown that learning unsupervised topics in this fashion maximizes the total information gained, about all texts $\mathcal{D}$ when described by all words $\mathcal{W}$.

    \[\mathcal{I}(\mathcal{D}, \mathcal{W})=\sum_{d \in \mathcal{D}} \sum_{w \in \mathcal{W}} P(d, w) \log \left(\frac{P(d, w)}{P(d) P(w)}\right)\]

This characteristic is handy for finding relevant samples across varying topics, allowing us to search within the learned documents $d_i$. We train a deep topic model $\mathcal{M}_\textit{topic}$ using multi-domain data $\mathcal{D}$ and obtain topic clusters $\mathcal{C} = (\mathcal{C}_i, \dots \mathcal{C}_t)$, where $\lvert \mathcal{C} \rvert = t$ is the number of topic clusters. We obtain the vector representation for $\forall d_i$ from the tuned PLM embeddings $\mathcal{E} = (e_1, \dots e_m)$ in $\mathcal{M}_\textit{topic}$, while iteratively traversing through the clusters $\mathcal{C}_i \in \mathcal{C}$.

\begin{algorithm}
\caption{Topic Efficient Sampling}\label{algo:sampling}
\begin{algorithmic}
\Require $S \geq 0$ \Comment{Sampling Threshold} 
\Require $\textit{Avg} \in \{\textit{moving}, \textit{exp} \}$
\Ensure $\lvert \mathcal{C} \rvert > 0$
\State $\mathcal{D}_{\textbf{train}} \gets \{\}$
\State $I \gets \{\frac{\lvert \mathcal{C}_1 \rvert}{\sum \limits_{\mathcal{C}_i \in \mathcal{C}} \mathcal{C}_i} \dots \frac{\lvert \mathcal{C}_t \rvert}{\sum \limits_{\mathcal{C}_i \in \mathcal{C}} \mathcal{C}_i} \}$ \Comment{Cluster Importances}
\For{$\mathcal{C}_i \in \mathcal{C}$} \Comment{Iterating for each cluster}
\State $\mathcal{E}_i \gets \{\textit{PLM}(d_i^1) \dots \} = \{\mathbf{e}_i^1 \dots \mathbf{e}_i^m \}$
\State $s_i \gets max(1,S \cdot I_i)$ \Comment{Threshold per cluster}
\State $\textit{j} \gets 0$
\State $\textbf{\textit{cent}}_0 \gets \frac{\sum \limits_{\mathbf{e}_i \in \mathcal{E}} \mathbf{e}_i}{\lvert \mathcal{E} \rvert}$ \Comment{Centroid of the cluster}
\While{$\textit{j} \leq s_i$}
\State $\textbf{\textit{sim}} = \frac{\langle\mathcal{E}, \textit{cent}\rangle}{\|\mathbf{\mathcal{E}}\|\|\textit{cent}\|}$ \Comment{Similarity Ranking}
\State $\textit{sample}=\argsort (\textbf{\textit{sim}}, \textbf{Ascending})[0]$
\Comment{Take the sample most diverse from the centroid}
\State $\mathcal{D}_\textbf{train} \gets \mathcal{D}_\textbf{train} \cup \textit{sample}$
\State $\textit{j} \gets \textit{j} + 1$
\State \resizebox{0.8\hsize}{!}{$ \textbf{\textit{cent}}_j \gets \begin{cases}  \alpha \cdot \mathbf{e}_\textit{sample}+(1-\alpha)\cdot \textit{cent}_{j-1} & \textit{exp}\\ \frac{(j-1)}{j} \cdot
\textbf{\textit{cent}}_j+ \frac{\mathbf{e}_\textit{sample}}{j} & \textit{moving}
\end{cases}$}

\State \Comment{Centroid update w.r.t. sampled data}

\EndWhile{}
\EndFor{}\\
\Return $\mathcal{D}_\textbf{train}$
\end{algorithmic}
\end{algorithm}

Our sampling process selects increasingly more diverse samples after each iteration. This search within the relevant examples is presented in \autoref{algo:sampling}.
This algorithm selects a set of diverse samples from the given multi-domain datasets $\mathcal{D}$, using the clusters from a deep topic model $\mathcal{M}_\textit{topic}$ and the sentence embeddings $\mathcal{E}$ of the sentences as a basis for comparison. The algorithm starts by selecting a random sentence as the first diverse sample and uses this sentence to calculate a ``centroid'' embedding. It then iteratively selects the next most dissimilar sentence to the current centroid, until the desired number of diverse samples is obtained.


\subsection{Topic-Guided Stance Detection}\label{sec:methods:stance}

\paragraph{Task Formalization}

Given the topic, $t_i$ for each document $d_i$ in the generated set $\mathcal{D}_\textbf{train}$ we aim to classify the stance expressed within that text towards the topic. For a fair comparison with prior work, we use the label mapping from the previous multi-domain benchmark \citep{hardalov2021cross} and standardise the original labels $L$ into a five-way stance classification setting, $S = \{\text{Positive, Negative, Discuss, Other, Neutral}\}$. Stance detection can be generalized as pairwise sequence classification, where a model learns a mapping $f:(d_i, t_i) \to S$. We combine the textual sequences with the stance labels to learn this mapping. 
The combination is implemented using a simple prompt commonly used for NLI tasks \cite{lan2019albert, raffel2020exploring,hambardzumyan2021warp}, where the textual sequence becomes the premise and the topic the hypothesis.

\begin{equation*} \label{prompt:combination}
\begin{split}
& \text{[CLS] premise: \textit{premise}} \\ 
& \text{hypothesis: \textit{topic} [EOS]}
\end{split}
\end{equation*}

The result of this process is a supervised dataset for stance prediction $\mathcal{D}_\textbf{train} = ((Prompt(d_1,t_1), s_1) \dots (Prompt(d_n,t_n), s_n))$ where $\forall s_i \in S$. This method allows for data-efficient sampling, as we at most sample $10\%$ of the data while preserving the diversity and relevance of the selected samples. The versatility of the method allows \emph{TESTED} to be applied to any text classification setting.

\paragraph{Tuning with a Contrastive Objective}

After obtaining the multi-domain supervised training set $\mathcal{D}_\textbf{train}$, we decided to leverage the robustness of PLMs, based on a transformer architecture \cite{vaswani2017attention} and fine-tune on $\mathcal{D}_\textbf{train}$ with a single classification head. This effectively allows us to transfer the knowledge embedded within the PLM onto our problem domain. For standard fine-tuning of the stance detection model $\mathcal{M}_\textit{stance}$ we use cross-entropy as our initial loss:
\begin{equation}
\mathcal{L}_\textit{CE} = -\sum \limits_{i \in S}  y_i \log \left(\mathcal{M}_\textit{stance}(d_i)\right)
\end{equation}

Here $y_i$ is the ground truth label. However, as we operate in a multi-domain setting, with variations in writing vocabulary, style and covered topics, it is necessary to train a model where similar sentences have a homogeneous representation within the embedding space while keeping contrastive pairs distant. We propose a new contrastive objective based on the \textit{cosine} distance between the samples to accomplish this. In each training batch $B=(d_1, \dots d_b)$, we create a matrix of contrastive pairs $\mathcal{P} \in \mathcal{R}^{b \times b}$, where $\forall i,j =\overline{1,b}$, $\mathcal{P}_{ij} =1$ if $i$-th and $j$-th examples share the same label and $-1$ otherwise. The matrices can be precomputed during dataset creation, thus not adding to the computational complexity of the training process. We formulate our pairwise contrastive objective $\mathcal{L}_\textit{CL}(x_i, x_j, \mathcal{P}_{ij})$ using matrix $\mathcal{P}$.

\begin{align}
\mathcal{L}_\textit{CL}=\resizebox{.72\hsize}{!}{$ \begin{cases} e(1- e^{\cos \left(x_i, x_j\right) - 1)}, \mathcal{P}_{ij}=1 \\ e^{\max \left(0, \cos \left(x_i, x_j\right)-\beta\right)}-1, \mathcal{P}_{ij}=-1\end{cases}$
}
\end{align}

Here $x_i,x_j$ are the vector representations of examples $d_i,d_j$. The loss is similar to cosine embedding loss and soft triplet loss
 \citep{barz2020deep,qian2019softtriple}; however, it penalizes the opposing pairs harsher because of the exponential nature, but does not suffer from computational instability as the values are bounded in the range $[0,e - \frac{1}{e}]$. 
 The final loss is:  
 \begin{align}
     \mathcal{L} = \mathcal{L}_\textit{CE} + \mathcal{L}_\textit{CL}
 \end{align}

We use the fine-tuning method from \citet{mosbach2020stability, liu2019roberta} to avoid the instability caused by catastrophic forgetting, small-sized fine-tuning datasets or optimization difficulties.

\section{Experimental Setup}
\label{sec:experiments}
\setlength{\tabcolsep}{3pt}
\begin{table*}
\centering
\begin{adjustbox}{max width=\textwidth}
\begin{tabular}{@{}lc|ccccc|ccc|ccccc|ccc@{}}
\toprule
 &
  $\mathrm{F}_1$ avg. &
  \rotatebox{45}{arc} &
  \rotatebox{45}{iac1} &
  \rotatebox{45}{perspectrum} &
  \rotatebox{45}{poldeb} &
  \rotatebox{45}{scd} &
  \rotatebox{45}{emergent} &
  \rotatebox{45}{fnc1} &
  \rotatebox{45}{snopes} &
  \rotatebox{45}{mtsd} &
  \rotatebox{45}{rumor} &
  \rotatebox{45}{semeval16} &
  \rotatebox{45}{semeval19} &
  \rotatebox{45}{wtwt} &
  \rotatebox{45}{argmin} &
  \rotatebox{45}{ibmcs} &
  \rotatebox{45}{vast} \\ \midrule
Majority class baseline &
  27.60 &
  21.45 &
  21.27 &
  34.66 &
  39.38 &
  35.30 &
  21.30 &
  20.96 &
  43.98 &
  19.49 &
  25.15 &
  24.27 &
  22.34 &
  15.91 &
  33.83 &
  34.06 &
  17.19 \\
Random baseline &
  35.19 &
  18.50 &
  30.66 &
  50.06 &
  48.67 &
  50.08 &
  31.83 &
  18.64 &
  45.49 &
  33.15 &
  20.43 &
  31.11 &
  17.02 &
  20.01 &
  49.94 &
  50.08 &
  33.25 \\
MoLE &
  65.55 &
  63.17 &
  38.50 &
  85.27 &
  50.76 &
  \textbf{65.91} &
  \textbf{83.74} &
  75.82 &
  75.07 &
  \textbf{65.08} &
  \textbf{67.24} &
  \textbf{70.05} &
  57.78 &
  68.37 &
  \textbf{63.73} &
  79.38 &
  38.92 \\ \midrule
TESTED (Our Model) &
  \textbf{69.12} &
  \textbf{64.82} &
  \textbf{56.97} &
  \textbf{83.11} &
  \textbf{52.76} &
  64.71 &
  82.10 &
  \textbf{83.17} &
  \textbf{78.61} &
  63.96 &
  66.58 &
  69.91 &
  \textbf{58.72} &
  \textbf{70.98} &
  62.79 &
  \textbf{88.06} &
  \textbf{57.47} \\
Topic $\rightarrow$ Random Sampling &
  61.14 &
  53.92 &
  42.59 &
  77.68 &
  44.08 &
  52.54 &
  67.55 &
  75.60 &
  72.67 &
  56.35 &
  59.08 &
  66.88 &
  57.28 &
  69.32 &
  52.02 &
  76.93 &
  53.80 \\
Topic $\rightarrow$ Stratified Sampling &
  64.01 &
  50.27 &
  51.57 &
  77.78 &
  46.67 &
  62.13 &
  79.00 &
  77.90 &
  76.44 &
  61.50 &
  64.92 &
  68.45 &
  51.96 &
  69.47 &
  56.76 &
  78.30 &
  51.16 \\
- Contrastive Objective &
  65.63 &
  61.11 &
  55.50 &
  81.85 &
  43.81 &
  63.04 &
  80.84 &
  79.05 &
  73.43 &
  62.18 &
  61.57 &
  60.17 &
  56.06 &
  68.79 &
  59.51 &
  86.94 &
  56.35 \\
\begin{tabular}[c]{@{}l@{}}Topic Sampling $\rightarrow$ Stratified \\ - Contrastive Loss\end{tabular} &
  63.24 &
  60.98 &
  49.17 &
  77.85 &
  45.54 &
  58.23 &
  77.36 &
  75.80 &
  74.77 &
  60.85 &
  63.69 &
  62.59 &
  54.74 &
  62.85 &
  53.67 &
  86.04 &
  47.72 \\ \bottomrule
\end{tabular}%
\end{adjustbox}
\caption{In-domain results reported with macro averaged F1, averaged over experiments. In lines under \textit{TESTED}, we replace (for Sampling) $(\rightarrow)$ or remove (for loss) $(-)$, the comprising components. }
\label{tab:in_domiain_results}
\end{table*}
\begin{table*}
\centering
\begin{adjustbox}{max width=\textwidth}
\begin{tabular}{@{}lc|ccccc|ccc|ccccc|ccc@{}}
\toprule
 &
  $\mathrm{F_1}$ avg. &
  \rotatebox{45}{arc} &
  \rotatebox{45}{iac1} &
  \rotatebox{45}{perspectrum} &
  \rotatebox{45}{poldeb} &
  \rotatebox{45}{scd} &
  \rotatebox{45}{emergent} &
  \rotatebox{45}{fnc1} &
  \rotatebox{45}{snopes} &
  \rotatebox{45}{mtsd} &
  \rotatebox{45}{rumor} &
  \rotatebox{45}{semeval16} &
  \rotatebox{45}{semeval19} &
  \rotatebox{45}{wtwt} &
  \rotatebox{45}{argmin} &
  \rotatebox{45}{ibmcs} &
  \rotatebox{45}{vast} \\ \midrule
MoLE w/ Hard Mapping &
  32.78 &
  25.29 &
  35.15 &
  29.55 &
  22.80 &
  16.13 &
  58.49 &
  47.05 &
  29.28 &
  23.34 &
  32.93 &
  37.01 &
  21.85 &
  16.10 &
  34.16 &
  72.93 &
  22.89 \\
MoLE w/ Weak Mapping &
  49.20 &
  \textbf{51.81} &
  38.97 &
  58.48 &
  47.23 &
  53.96 &
  \textbf{82.07} &
  51.57 &
  56.97 &
  40.13 &
  \textbf{51.29} &
  36.31 &
  31.75 &
  22.75 &
  50.71 &
  75.69 &
  37.15 \\
MoLE w/Soft Mapping &
  46.56 &
  48.31 &
  32.21 &
  62.73 &
  54.19 &
  51.97 &
  46.86 &
  57.31 &
  53.58 &
  37.88 &
  44.46 &
  36.77 &
  28.92 &
  28.97 &
  57.78 &
  72.11 &
  30.96 \\ \midrule
TESTED &
  \textbf{59.41} &
  50.80 &
  \textbf{57.95} &
  \textbf{78.95} &
  \textbf{55.62} &
  \textbf{55.23} &
  80.80 &
  \textbf{72.51} &
  \textbf{61.70} &
  \textbf{55.49} &
  39.44 &
  \textbf{40.54} &
  \textbf{46.28} &
  \textbf{42.77} &
  \textbf{72.07} &
  \textbf{86.19} &
  \textbf{54.33} \\ \midrule
Topic Sampling $\rightarrow$ Stratified &
  50.38 &
  38.47 &
  46.54 &
  69.75 &
  50.54 &
  51.37 &
  68.25 &
  59.41 &
  51.64 &
  48.24 &
  28.04 &
  29.69 &
  34.97 &
  38.13 &
  63.83 &
  83.20 &
  44.06 \\
- Contrastive Loss &
  54.63 &
  47.96 &
  50.09 &
  76.51 &
  47.49 &
  51.93 &
  75.22 &
  68.69 &
  56.53 &
  49.47 &
  33.95 &
  37.96 &
  44.10 &
  39.56 &
  63.09 &
  83.59 &
  48.03 \\ \bottomrule
\end{tabular}%
\end{adjustbox}
\caption{Out-of-domain results with macro averaged F1. In lines under \textit{TESTED}, we replace (for Sampling) $(\rightarrow)$ or remove (for loss) $(-)$, the comprising components. Results for MoLE w/Soft Mapping are aggregated across with best per-embedding results present in the study \citep{hardalov2021cross}.}
\label{tab:ood_results}
\end{table*}
\setlength{\tabcolsep}{6pt}

\subsection{Evaluation}
We evaluate our method on the $16$ dataset multi-domain benchmark and the baselines proposed by \citet{hardalov2021cross}. To directly compare with prior work, we use the same set of evaluation metrics: macro averaged F1, precision, recall and accuracy.

\subsection{Model Details}

We explore several PLM transformer architectures within our training and classification pipelines in order to evaluate the stability of the proposed technique. We opt to finetune a pre-trained \textit{roberta-large} architecture \cite{liu2019roberta, conneau2019unsupervised}. For fine-tuning, we use the method introduced by \citet{mosbach2020stability}, by adding a linear warmup on the initial $10\%$ of the iteration raising the learning rate to $2e^{-5}$ and decreasing it to $0$ afterwards. We use a weight decay of $\lambda = 0.01$ and train for $3$ epochs with global gradient clipping on the stance detection task. We further show that learning for longer epochs does not yield sizeable improvement over the initial fine-tuning. The optimizer used for experimentation is an AdamW \citep{loshchilov2017decoupled} with a bias correction component added to stabilise the experimentation \citep{mosbach2020stability}. 

\paragraph{Topic Efficiency}

Recall that we introduce a topic-guided diversity sampling method within \textbf{\textit{TESTED}}, which allows us to pick relevant samples per topic and class for further fine-tuning. We evaluate its effectiveness by fine-tuning PLMs on the examples it generates and comparing it with training on a random stratified sample of the same size. 

\section{Results and Analysis}

In this section, we discuss and analyze our results, while comparing the performance of the method against the current state-of-the-art \citep{hardalov2021cross} and providing an analysis of the topic efficient sampling and the contrastive objective.

\subsection{Stance Detection}

\paragraph{In-domain}

We train on our topic-efficient subset $\mathcal{D}_\textbf{train}$ and test the method on all datasets $\mathcal{D}$ in the multi-domain benchmark. Our method TESTED is compared to MoLE \citep{hardalov2021cross}, a strong baseline and the current state-of-the-art on the benchmark. The results, presented in \autoref{tab:in_domiain_results}, show that TESTED has the highest average performance on in-domain experiments with an increase of $3.5$ F1 points over MoLE, all while using $\leq10\%$ of the amount of training data in our subset $\mathcal{D}_\textbf{train}$ sampled from the whole dataset $\mathcal{D}$. Our method is able to outperform all the baselines on $10$ out of $16$ datasets. On the remaining $6$ datasets the maximum absolute difference between TESTED and MoLE is $1.1$ points in F1. 
We also present ablations for TESTED, by replacing the proposed sampling method with other alternatives, removing the contrastive objective or both simultaneously. Replacing Topic Efficient sampling with either \emph{Random} or \emph{Stratified} selections deteriorates the results for all datasets with an average decrease of $8$ and $5$ F1 points, respectively. We attribute this to the inability of other sampling techniques to maintain inter-topic distribution and per-topic label distributions balanced while selecting diverse samples. We further analyse how our sampling technique tackles these tasks in \autoref{subsec:imbalance}. We also see that removing the contrastive loss also results in a deteriorated performance across all the datasets with an average decrease of $3$ F1 points. In particular, we see a more significant decrease in datasets with similar topics and textual expressions, i.e. \emph{poldeb} and \emph{semeval16}, meaning that learning to differentiate between contrastive pairs is essential within this task. We analyse the effect of the contrastive training objective further in \autoref{subsec:contrastive}.


\paragraph{Out-of-domain}

In the out-of-domain evaluation, we leave one dataset out of the training process for subsequent testing. We present the results of TESTED in \autoref{tab:ood_results}, showing that it is able to overperform over the previous state-of-the-art significantly. The metrics in each column of \autoref{tab:ood_results} show the results for each dataset held out from training and only evaluated on. Our method records an increased performance on $13$ of $16$ datasets, with an averaged increase of $10.2$ F1 points over MoLE, which is a significantly more pronounced increase than for the in-domain setting, demonstrating that the strength of TESTED lies in better out-of-domain generalisation. We can also confirm that replacing the sampling technique or removing the contrastive loss results in lower performance across all datasets, with decreases of $9$ and $5$ F1 points respectively. This effect is even more pronounced compared to the in-domain experiments, as adapting to unseen domains and topics is facilitated by diverse samples with a balanced label distribution.

\subsection{Imbalance Mitigation Through Sampling}
\label{subsec:imbalance}
\begin{figure*}
\centering
\includegraphics[width=\textwidth]{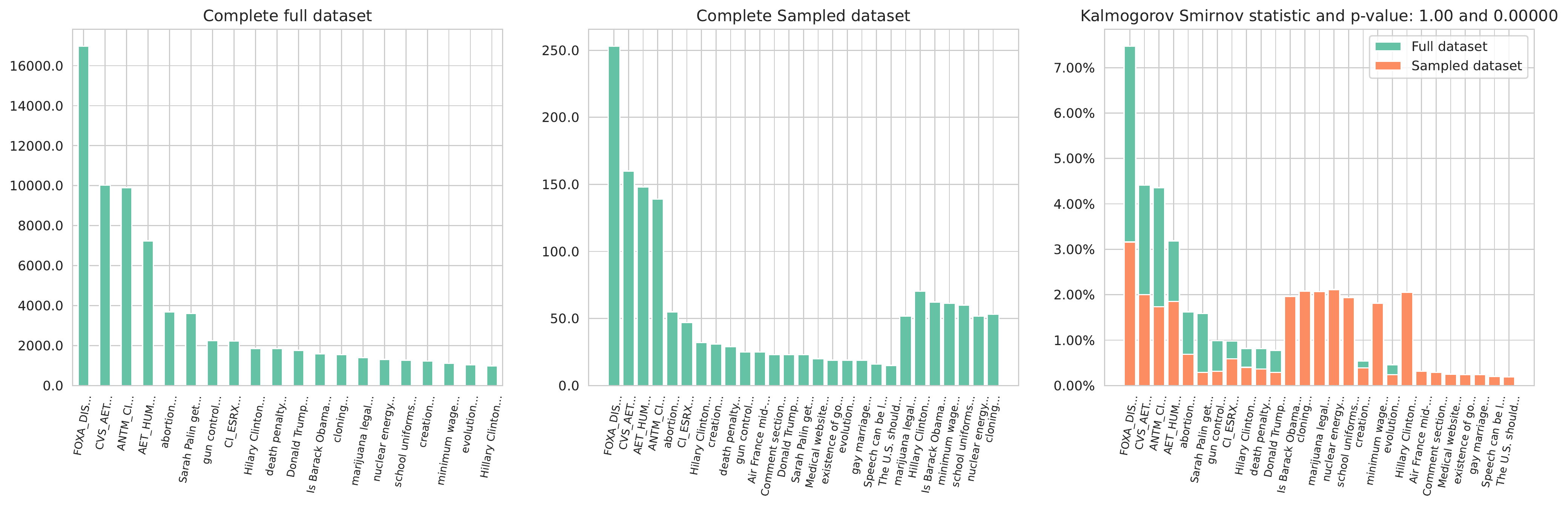}
\caption{Distributions of top 20 most frequent topics in complete dataset $\mathcal{D}$ (left), Sampled dataset $\mathcal{D}_\textbf{train}$ (mid) and their aggregated comparison (right). The distribution of top $20$ topics in $\{\mathcal{D}\} - \{\mathcal{D}_\textbf{train}\}$ is added to the tail of the figure (mid). 
}
\label{fig:imbalanced}
\end{figure*}

\paragraph{Inter-Topic}
To investigate the inter-topic imbalances, we look at the topic distribution for the top $20$ most frequent topics covered in the complete multi-domain dataset $\mathcal{D}$, which accounts for $\geq 40 \%$ of the overall data. As we can see in  \autoref{fig:imbalanced}, even the most frequent topics greatly vary in their representation frequency, with $\sigma = 4093.55$, where $\sigma$ is the standard deviation between represented amounts. For the training dataset $\mathcal{D}_\textbf{train}$, by contrast, the standard deviation between the topics is much smaller $\sigma = 63.59$. This can be attributed to the fact that $\mathcal{D}_\textbf{train}$ constitutes  $\leq10\%$ of $\mathcal{D}$, thus we also show the aggregated data distributions in \autoref{fig:imbalanced}. 
For a more systematic analysis, we employ the two sample Kolmogorov-Smirnov (KS) test \citep{kalmagorov}, to compare topic distributions in $\mathcal{D}$ and $\mathcal{D}_\textbf{train}$  for each dataset present in $\mathcal{D}$. The test compares the cumulative distributions (CDF) of the two groups, in terms of their maximum-absolute difference, $\text { stat }=\sup _x\left|F_1(x)-F_2(x)\right|$.

\begin{table}
\centering
\fontsize{10}{10}\selectfont
\begin{tabular}{@{}lrr@{}}
\toprule
 dataset                    & stat  & p-value  \\ \midrule
\textbf{fnc-1-ours}         & 1.00 & 0.007937 \\
\textbf{arc}                & 0.40 & 0.873016 \\
\textbf{emergent}           & 0.80 & 0.079365 \\
wtwt                        & 0.20 & 1.000000 \\
\textbf{rumor}              & 0.40 & 0.873016 \\
\textbf{snopes}             & 0.40 & 0.873016 \\
\textbf{perspectrum}        & 0.60 & 0.357143 \\
\textbf{vast}               & 0.60 & 0.357143 \\
\textbf{semeval2016task6}   & 0.40 & 0.873016 \\
\textbf{iac}                & 0.40 & 0.873016 \\
mtsd                        & 0.25 & 1.000000 \\
\textbf{argmin}             & 0.40 & 0.873016 \\
\textbf{scd}                & 1.00 & 0.007937 \\
\textbf{ibm\_claim\_stance} & 0.80 & 0.079365 \\
\textbf{politicaldebates}   & 0.50 & 1.000000 \\ \bottomrule
\end{tabular}%
\caption{KS test for topic distributions. The topics in bold designate a rejected null-hypothesis (criteria: $p\leq0.05$ or $\textit{stat} \geq 0.4$), that the topics in $\mathcal{D}$ and $\mathcal{D}_\textbf{train}$ come from the same distribution.}
\label{tab:kalmagorov}
\end{table}

The results in \autoref{tab:kalmagorov} show that the topic distribution within the full and sampled data $\mathcal{D}$, $\mathcal{D}_\textbf{train}$, cannot be the same for most of the datasets. The results for the maximum-absolute difference also show that with at least $0.4$ difference in CDF, the sampled dataset $\mathcal{D}_\textbf{train}$ on average has a more balanced topic distribution. The analysis in \autoref{fig:imbalanced} and \autoref{tab:kalmagorov}, show that the sampling technique is able to mitigate the inter-topic imbalances present in $\mathcal{D}$. A more in-depth analysis for each dataset is provided in \autoref{Appendix:imbalances}.

\paragraph{Per-topic}

\begin{figure}
\centering
\includegraphics[width=\columnwidth]{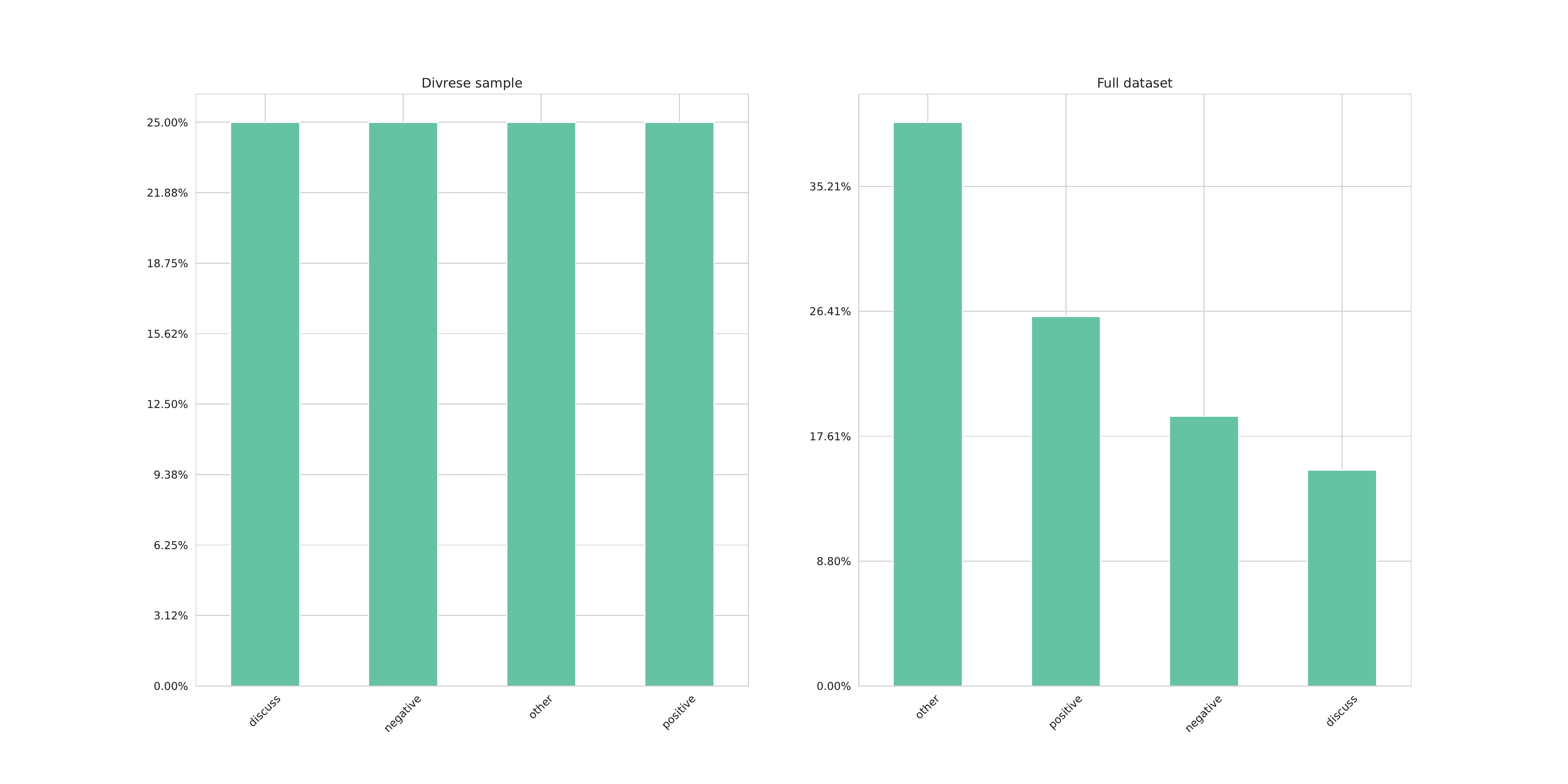}
\caption{Label distribution in $\mathcal{D}$ (right) and $\mathcal{D}_\textbf{train}$ (left).}
\label{fig:label_dist}
\end{figure}

For the per-topic imbalance analysis, we complete similar steps to the inter-topic analysis, with the difference that we iterate over the top $20$ frequent topics looking at \emph{label} imbalances within each topic. We examine the label distribution for the top $20$ topics for a per-topic comparison. The standard deviation in label distributions averaged across those 20 topics is $\sigma=591.05$ for the whole dataset $\mathcal{D}$ and the sampled set $\mathcal{D}_\textbf{train}$ $\sigma=11.7$. This can be attributed to the stratified manner of our sampling technique. This is also evident from \autoref{fig:label_dist}, which portrays the overall label distribution in $\mathcal{D}$ and $\mathcal{D}_\textbf{train}$. 

To investigate the difference in label distribution for each of the top 20 topics in $\mathcal{D}$, we use the KS test, presented in \autoref{tab:per_topic_imbalance}. For most topics, we see that the label samples in $\mathcal{D}$ and $\mathcal{D}_\textbf{train}$ cannot come from the same distribution. This means that the per-topic label distribution in the sampled dataset $\mathcal{D}_\textbf{train}$, does not possess the same imbalances present in $\mathcal{D}$.

\begin{table}[t]
\centering
\fontsize{8}{9}\selectfont
\begin{tabular}{@{}lr@{}}
\toprule
topic                                  & p-values \\ \midrule
\textbf{FOXA\_DIS}                     & 0.028571 \\
\textbf{CVS\_AET}                      & 0.028571 \\
\textbf{ANTM\_CI}                      & 0.028571 \\
\textbf{AET\_HUM}                      & 0.047143 \\
abortion                               & 0.100000 \\
\textbf{Sarah Palin getting divorced?} & 0.028571 \\
\textbf{gun control}                   & 0.001879 \\
\textbf{CI\_ESRX}                      & 0.028571 \\
\textbf{Hilary Clinton}                & 0.001468 \\
death penalty                          & 0.100000 \\
\textbf{Donald Trump}                  & 0.002494 \\
\textbf{Is Barack Obama muslim?}       & 0.028571 \\
cloning                                & 0.333333 \\
\textbf{marijuana legalization}        & 0.032178 \\
nuclear energy                         & 0.333333 \\
school uniforms                        & 0.333333 \\
\textbf{creation}                      & 0.003333 \\
minimum wage                           & 0.333333 \\
evolution                              & 0.100000 \\
\textbf{lockdowns}                     & 0.000491 \\ \bottomrule
\end{tabular}
\caption{KS test for label distributions. The topics in bold designate a rejected null-hypothesis (criteria: $p\leq0.05$), that the label samples in $\mathcal{D}$ and $\mathcal{D}_\textbf{train}$ averaged per top $20$ topics come from the same distribution.}
\label{tab:per_topic_imbalance}
\end{table}

\begin{figure}[t]
\centering
\includegraphics[width=\columnwidth]{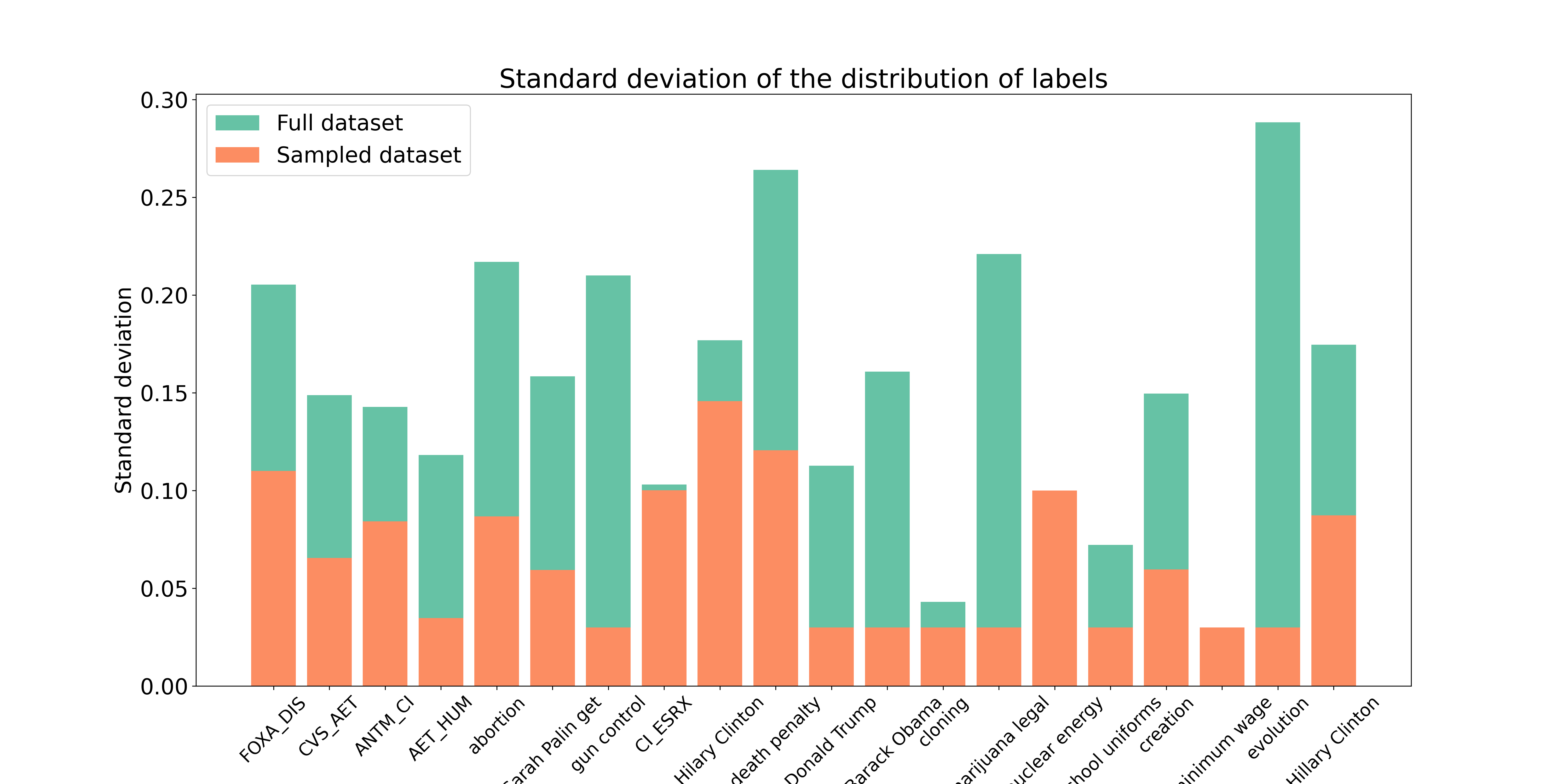}
\caption{Normalized Standard Deviation in label distribution for top 20 topics.
}
\label{fig:per_topic_dist_std}
\end{figure}

We can also see the normalized standard deviation for the label distribution within $\mathcal{D}_\textbf{train}$ is lower than in $\mathcal{D}$, as shown in \autoref{fig:per_topic_dist_std}. This reinforces the finding that per-topic label distributions in the sampled dataset are more uniform. For complete per-topic results, we refer the reader to  \autoref{Appendix:imbalances}.

\paragraph{Performance} 

Using our topic-efficient sampling method is highly beneficial for in- and out-of-domain experiments, presented in \autoref{tab:in_domiain_results} and \autoref{tab:ood_results}. Our sampling method can select diverse and representative examples while outperforming \textit{Random} and \textit{Stratified} sampling techniques by $8$ and $5$ F1 points on average. This performance can be attributed to the mitigated inter- and per-topic imbalance in $\mathcal{D}_\textbf{train}$.

\subsection{Data Efficiency}
\label{subsec:data_efficiency}

\begin{figure}[t]
\centering
\includegraphics[width=\columnwidth]{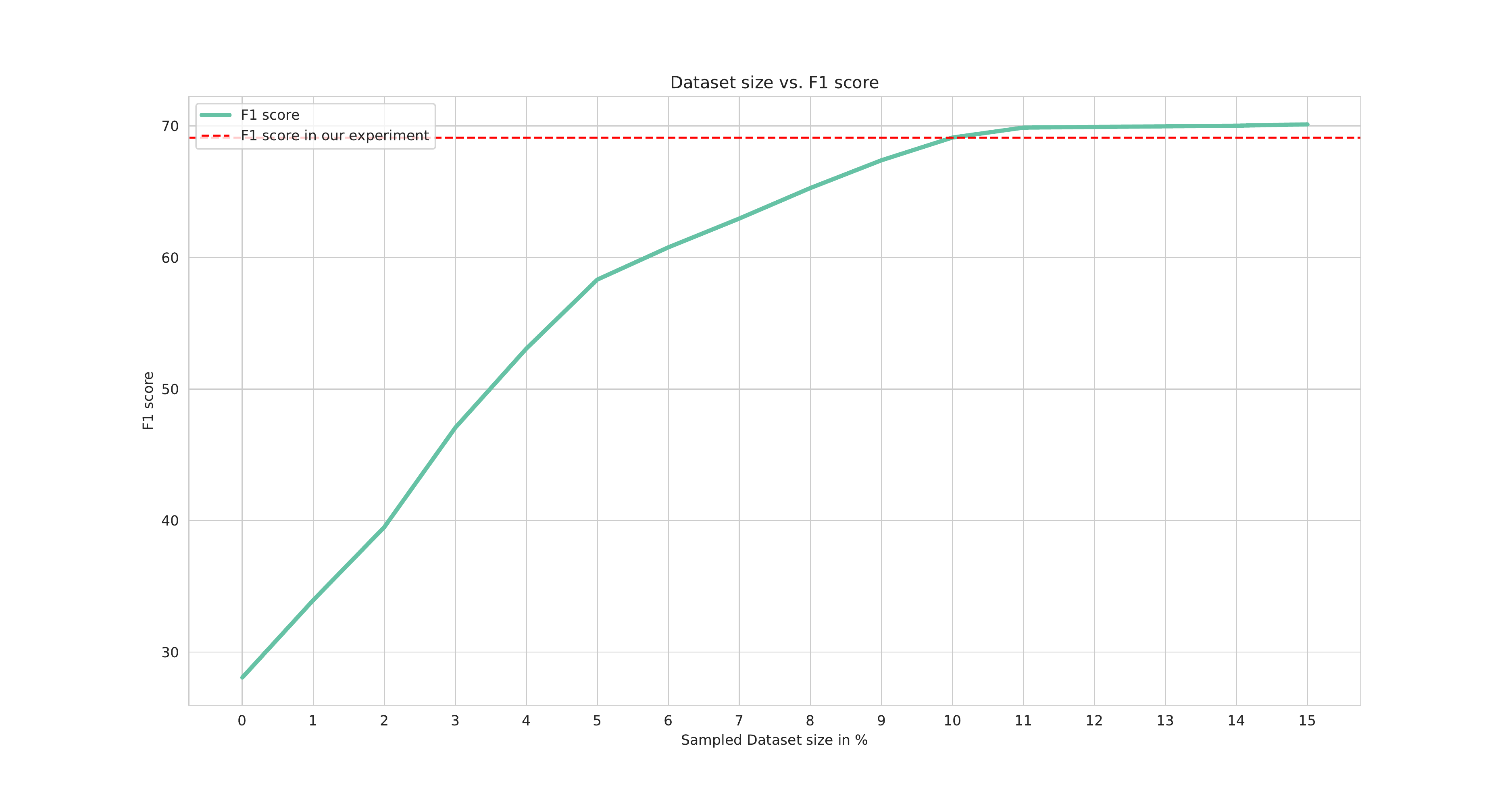}
\caption{Sampled Data size vs Performance. Performance increases with a bigger sampled selection.}
\label{fig:data_efficient}
\end{figure}

TESTED allows for sampling topic-efficient, diverse and representative samples while preserving the balance of topics and labels. This enables the training of data-efficient models for stance detection while avoiding redundant or noisy samples. We analyse the data efficiency of our method by training on datasets with sizes $[1\%,15\%]$ compared to the overall data size $\lvert \mathcal{D}\rvert$, sampled using our technique. Results for the in-domain setting in terms of averaged F1 scores for each sampled dataset size are shown in \autoref{fig:data_efficient}. One can observe a steady performance increase with the more selected samples, but diminishing returns from the $10\%$ point onwards. This leads us to use $~10\%$ as the optimal threshold for our sampling process, reinforcing the data-efficient nature of TESTED.

\subsection{Contrastive Objective Analysis}
\label{subsec:contrastive}

\begin{figure}
\centering
\includegraphics[width=\columnwidth]{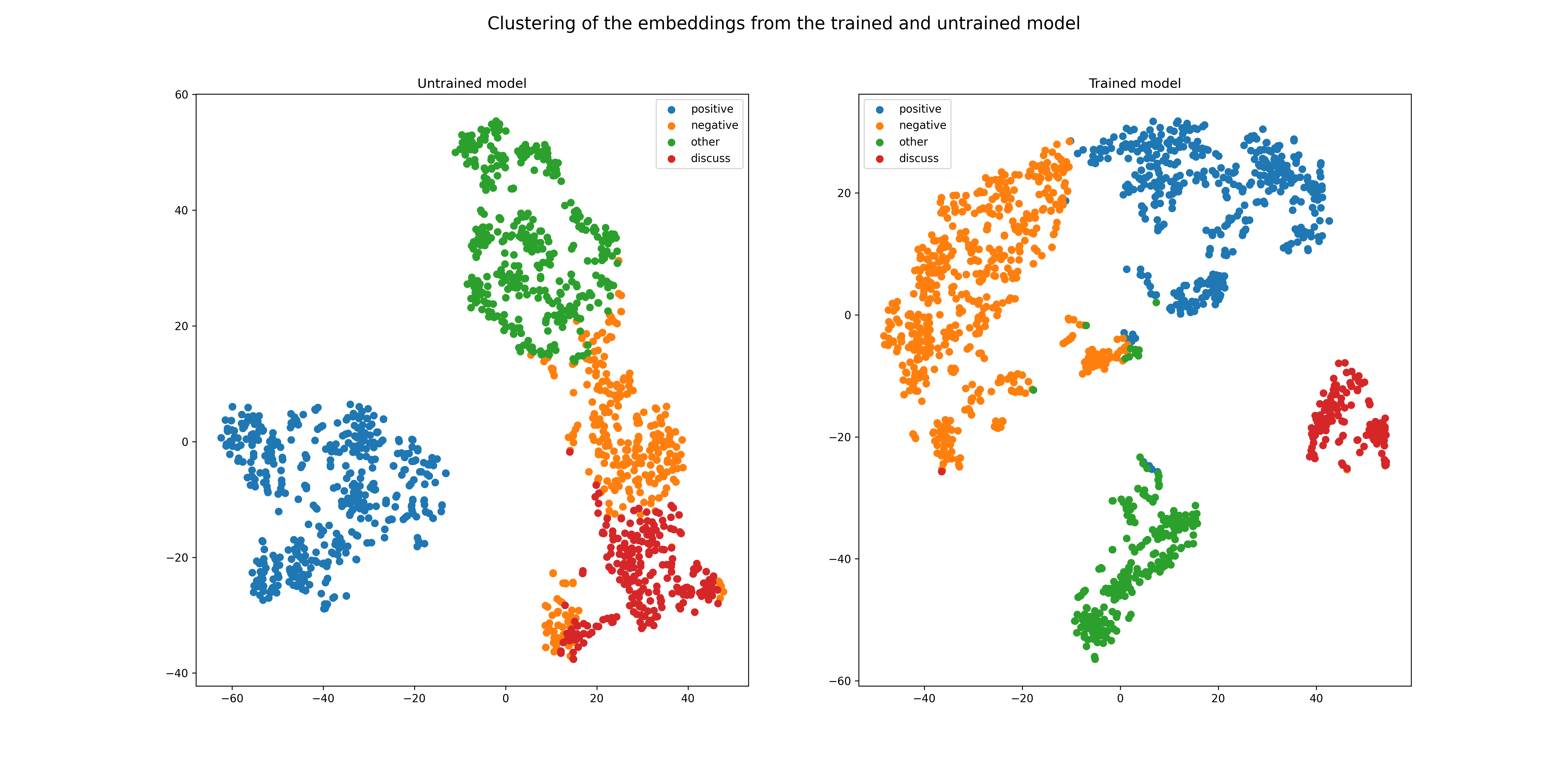}
\caption{Sample Representation before (left) and after (right) contrastive training.}
\label{fig:contrastive}
\end{figure}

To analyse the effect of the contrastive loss, we sample $200$ unseen instances stratified across each dataset and compare the sentence representations before and after training. To compare the representations, we reduce the dimension of the embeddings with t-SNE and cluster them with standard K-means. We see in \autoref{fig:contrastive} that using the objective allows for segmenting contrastive examples in a more pronounced way. The cluster purity also massively rises from $0.312$ to $0.776$ after training with the contrastive loss. This allows the stance detection model to differentiate and reason over the contrastive samples with greater confidence.

\section{Conclusions}
\label{sec:conclusions}

We proposed TESTED, a novel end-to-end framework for multi-domain stance detection. The method consists of a data-efficient topic-guided sampling module, that mitigates the imbalances inherent in the data while selecting diverse examples, and a stance detection model with a contrastive training objective. TESTED yields significant performance gains compared to strong baselines on in-domain experiments, but in particular generalises well on out-of-domain topics, achieving a $10.2$ F1 point improvement over the state of the art, all while using $\leq10\%$ of the training data. While in this paper, we have evaluated TESTED on stance detection, the method is applicable to text classification more broadly, which we plan to investigate in more depth in future work. 

\section*{Limitations}

Our framework currently only supports English, thus not allowing us to complete a cross-lingual study. Future work should focus on extending this study to a multilingual setup. Our method is evaluated on a $16$ dataset stance benchmark, where some domains bear similarities. The benchmark should be extended and analyzed further to find independent datasets with varying domains and minimal similarities, allowing for a more granular out-of-domain evaluation.

\label{sec:limitations}

\section*{Acknowledgements}

This research is funded by a DFF Sapere Aude research leader grant under grant agreement No 0171-00034B, as well as supported by the Pioneer Centre for AI, DNRF
grant number P1.

\bibliography{acl_latex}

\clearpage
\appendix

\section*{Appendix}
\label{sec:Appendix}


\section{Imbalance analysis}
\label{Appendix:imbalances}

\subsection{Inter-topic}
\label{Appendix:inter_topic_analysis}

\begin{figure*}
\centering
\includegraphics[width=\textwidth]{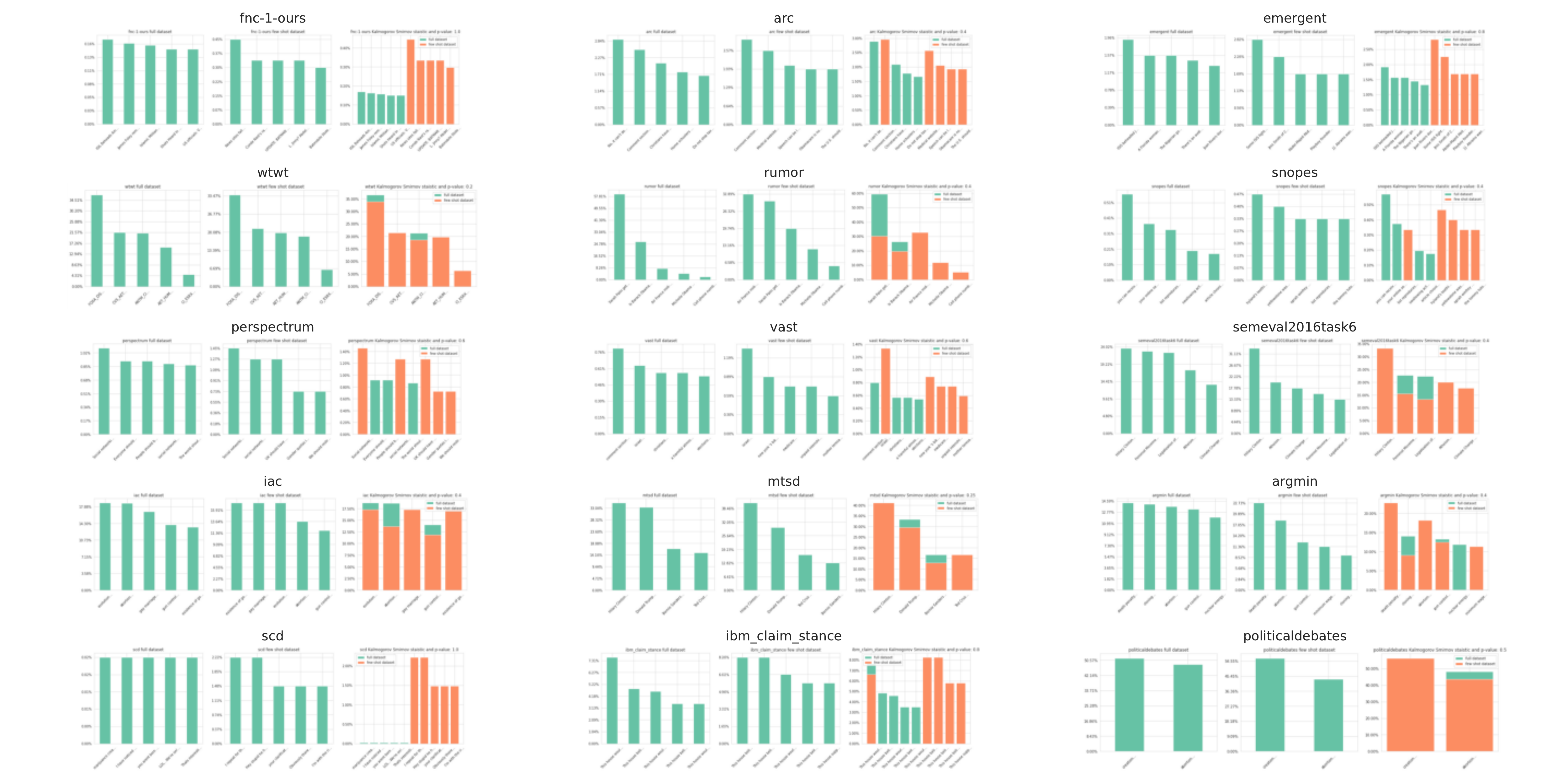}
\caption{Distributions of top 20 most frequent topics for each dataset (left), Sampled dataset $\mathcal{D}_{\textbf{train} = \textit{dataset}}$ (mid) and their aggregated comparison (right).
}
\label{fig:imbalanced_big}
\end{figure*}

To complement our inter-topic imbalance mitigation study, we complete an ablation on all topics in $\mathcal{D}$ and report them on a per-domain basis in \autoref{fig:imbalanced_big}. The trend is similar to the one in \autoref{fig:imbalanced}, where the dataset with imbalanced distributions is rebalanced, and balanced datasets are not corrupted.

\subsection{Per-topic}
\label{Appendix:per_topic_analysis}

\begin{figure*}
\centering
\includegraphics[width=\textwidth]{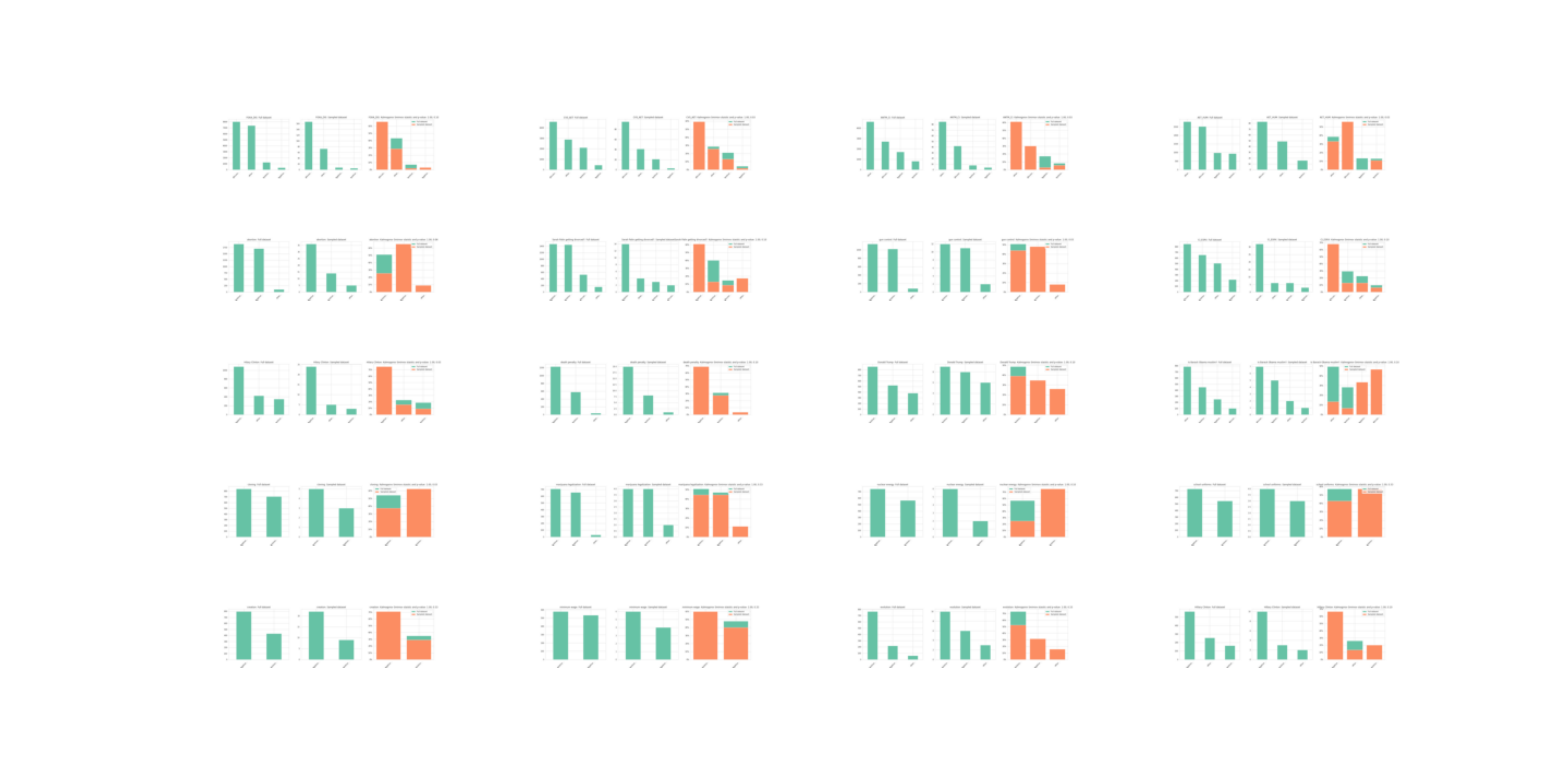}
\caption{Distributions of labels for top 20 most frequent topics for $\mathcal{D}$ (left), Sampled dataset $\mathcal{D}_{\textbf{train} = \textit{dataset}}$ (mid) and their aggregated comparison (right).
}
\label{fig:per_topic_big}
\end{figure*}

We show that our topic-efficient sampling method allows us to balance the label distribution for unbalanced topics, while not corrupting the ones distributed almost uniformly. To do this, we investigate each of the per-topic label distributions for the top $20$ most frequent topics while comparing  the label distributions for $\mathcal{\mathcal{D}}$ and $\mathcal{D}_\textbf{train}$, presented in \autoref{fig:per_topic_big}.

\section{Evaluation Metrics}

To evaluate our models and have a fair comparison with the introduced benchmarks we use a standard set of metrics for classification tasks such as macro-averaged F1, precision, recall and accuracy.
\begin{gather}
Acc = \frac{TP+TN}{TP+TN+FP+FN} \\
Prec = \frac{TP}{TP+FP} \\
Recall = \frac{TP}{TP+FN} \\
F1 = \frac{2*Prec*Recall}{Prec+Recall} = \frac{2*TP}{2*TP+FP+FN}
\end{gather}

\section{Dataset Statistics}
\label{Appendix:data}
We use a stance detection benchmark \citep{hardalov2021cross} whose data statistics are shown in Table \ref{tab:data_stat}. The label mapping employed is shown in Table \ref{tab:label_mapping}.

\begin{table}[t]
\centering
\resizebox{\columnwidth}{!}{
\begin{tabular}{lrrrr}
\toprule Dataset & Train & Dev & Test & Total \\
\midrule arc & 12,382 & 1,851 & 3,559 & 17,792 \\
argmin & 6,845 & 1,568 & 2,726 & 11,139 \\
emergent & 1,770 & 301 & 524 & 2,595 \\
fnc1 & 42,476 & 7,496 & 25,413 & 75,385 \\
iac1 & 4,227 & 454 & 924 & 5,605 \\
ibmcs & 935 & 104 & 1,355 & 2,394 \\
mtsd & 3,718 & 520 & 1,092 & $5,330$ \\
perspectrum & 6,978 & 2,071 & 2,773 & 11,822 \\
poldeb & 4,753 & 1,151 & 1,230 & 7,134 \\
rumor & 6,093 & 471 & 505 & $7,276$ \\
scd & 3,251 & 624 & 964 & 4,839 \\
semeval2016t6 & 2,497 & 417 & 1,249 & 4,163 \\
semeval2019t7 & 5,217 & 1,485 & 1,827 & 8,529 \\
snopes & 14,416 & 1,868 & 3,154 & 19,438 \\
vast & 13,477 & 2,062 & 3,006 & 18,545 \\
wtwt & 25,193 & 7,897 & 18,194 & 51,284 \\
\midrule Total & 154,228 & 30,547 & 68,495 & 253,270 \\
\bottomrule
\end{tabular}
}
\caption{Dataset statistics of the stance detection benchmark by \citet{hardalov2021cross} also used in this paper. Note that the rumour and mtsd datasets are altered in that benchmark as some of the data was unavailable.}
\label{tab:data_stat}
\end{table}

\begin{table}[t]
\centering
\resizebox{\columnwidth}{!}{%
\begin{tabular}{@{}ll@{}}
\toprule
Label    & Description                                                             \\ \midrule
Positive & agree, argument for, for, pro, favor, support, endorse                  \\
Negative & disagree, argument against, against, anti, con, undermine, deny, refute \\
Discuss  & discuss, observing, question, query, comment                            \\
Other    & unrelated, none, comment                                                \\
Neutral  & neutral                                                                 \\ \bottomrule
\end{tabular}%
}
\caption{Hard stance label mapping employed in this paper, following the stance detection benchmark by \citet{hardalov2021cross}.}
\label{tab:label_mapping}
\end{table}

\section{TESTED with different backbones}
\label{sec:transformers}

We chose to employ different PLM's as the backbone for TESTED and report the results in the \autoref{tab:plm_effect}. The PLMs are taken from the set of \textit{roberta-base, roberta-large, xlm-roberta-base, xlm-roberta-large. }The differences between models with a similar number of parameters are marginal. We can see a degradation of the F1 score between the \textit{base} and \textit{large}  versions of the models, which can be attributed to the expressiveness the models possess. We also experiment with the distilled version of the model and can confirm that in terms of the final F1 score, it works on par with the larger models. This shows that we can utilise smaller and more computationally efficient models within the task with marginal degradation in overall performance.

\setlength{\tabcolsep}{3pt}
\begin{table*}
\centering
\begin{adjustbox}{max width=\textwidth}
\begin{tabular}{@{}lc|ccccc|ccc|ccccc|ccc@{}}
\toprule
 &
  $\mathrm{F}_1$ avg. &
  \rotatebox{45}{arc} &
  \rotatebox{45}{iac1} &
  \rotatebox{45}{perspectrum} &
  \rotatebox{45}{poldeb} &
  \rotatebox{45}{scd} &
  \rotatebox{45}{emergent} &
  \rotatebox{45}{fnc1} &
  \rotatebox{45}{snopes} &
  \rotatebox{45}{mtsd} &
  \rotatebox{45}{rumor} &
  \rotatebox{45}{semeval16} &
  \rotatebox{45}{semeval19} &
  \rotatebox{45}{wtwt} &
  \rotatebox{45}{argmin} &
  \rotatebox{45}{ibmcs} &
  \rotatebox{45}{vast} \\ \midrule
  TESTED$_{\textit{reberta-large}}$ &
    69.12 &
    64.82 &
    56.97 &
    83.11 &
    52.76 &
    64.71 &
    82.10 &
    83.17 &
    78.61 &
    63.96 &
    66.58 &
    69.91 &
    58.72 &
    70.98 &
    62.79 &
    88.06 &
    57.47 \\
  TESTED$_{\textit{xlm-reberta-large}}$ &
    68.86 &
    64.35 &
    57.0 &
    82.71 &
    52.93 &
    64.75 &
    81.72 &
    82.71 &
    78.38 &
    63.66 &
    66.71 &
    69.76 &
    58.27 &
    71.29 &
    62.73 &
    87.75 &
    57.2 \\
  TESTED$_{\textit{reberta-base}}$ &
    65.32 &
    59.71 &
    51.86 &
    76.75 &
    50.23 &
    61.35 &
    78.84 &
    82.09 &
    73.31 &
    62.87 &
    65.46 &
    63.89 &
    58.3 &
    67.28 &
    58.28 &
    83.81 &
    51.09 \\
  TESTED$_{\textit{xlm-reberta-base}}$ &
    65.05 &
    60.26 &
    51.96 &
    76.2 &
    51.82 &
    58.74 &
    74.68 &
    77.9 &
    72.61 &
    62.71 &
    66.08 &
    69.74 &
    53.27 &
    65.83 &
    59.09 &
    87.92 &
    52.08 \\
    \midrule
  TESTED$_{\textit{distilroberta-base}}$ &
    68.86 &
    61.78 &
    56.94 &
    80.36 &
    46.29 &
    64.1 &
    79.26 &
    81.37 &
    73.44 &
    62.6 &
    63.4 &
    63.75 &
    56.53 &
    68.35 &
    57.27 &
    81.93 &
    56.3 \\
    \bottomrule
  \end{tabular}%
\end{adjustbox}
\caption{In-domain results reported with macro averaged F1, with varying backbones when using TESTED.}
\label{tab:plm_effect}
\end{table*}
\setlength{\tabcolsep}{6pt}

\end{document}